\documentclass[runningheads]{llncs}

\usepackage{eccv}

\usepackage{eccvabbrv}

\usepackage{graphicx}
\usepackage{booktabs}
\usepackage[utf8]{inputenc} 
\usepackage[T1]{fontenc}    
\usepackage{url}            
\usepackage{amsfonts}       
\usepackage{nicefrac}       
\usepackage{microtype}      
\usepackage{xcolor}         
\usepackage{algpseudocode}
\usepackage{multirow}
\usepackage{amsmath}
\usepackage{subcaption}
\usepackage{adjustbox}
\usepackage[outline]{contour}
\contourlength{0.2pt}
\contournumber{10}
\usepackage{enumitem}
\usepackage{cite}
\usepackage{amssymb}
\usepackage{algorithm}
\def\BibTeX{{\rm B\kern-.05em{\sc i\kern-.025em b}\kern-.08em
    T\kern-.1667em\lower.7ex\hbox{E}\kern-.125emX}}

\usepackage[accsupp]{axessibility}  

\usepackage[pagebackref,breaklinks,colorlinks,citecolor=eccvblue,hypertexnames=false]{hyperref}

\usepackage{orcidlink}

\newcommand{\missingcite}[1]{{\color{red} { [\textbf{CITATION NEEDED}]}}}

\DeclareMathAlphabet{\pazocal}{OMS}{zplm}{m}{n}

\begin{document}

\title{FlowPure: Continuous Normalizing Flows \\ for Adversarial Purification}



\makeatletter
\newcommand{\printfnsymbol}[1]{%
  \textsuperscript{\@fnsymbol{#1}}%
}
\makeatother

\author{
  Elias Collaert\thanks{Equal contribution.} \and Abel Rodríguez\printfnsymbol{1} \and Sander Joos\printfnsymbol{1} \and \\
  Lieven Desmet \and Vera Rimmer }
 

\authorrunning{E.~Collaert et al.}

\institute{DistriNet, KU Leuven \\
\email{\texttt{\{elias.collaert, abel.rodriguezromero, sander.joos} \\
\texttt{lieven.desmet, vera.rimmer\}@kuleuven.be}
}}


\maketitle

\begin{abstract}
  Despite significant advances in the area, adversarial robustness remains a critical challenge in systems employing machine learning models. The removal of adversarial perturbations at inference time, known as \textit{adversarial purification}, has emerged as a promising defense strategy. To achieve this, state-of-the-art methods leverage diffusion models that inject Gaussian noise during a forward process to dilute adversarial perturbations, followed by a denoising step to restore clean samples before classification. In this work, we propose \textbf{FlowPure}, a novel purification method based on Continuous Normalizing Flows (CNFs) trained with Conditional Flow Matching (CFM) to learn mappings from adversarial examples to their clean counterparts. Unlike prior diffusion-based approaches that rely on fixed noise processes, FlowPure can leverage specific attack knowledge to improve robustness under known threats, while also supporting a more general stochastic variant trained on Gaussian perturbations for settings where such knowledge is unavailable. Experiments on CIFAR-10 and CIFAR-100 demonstrate that our method outperforms state-of-the-art purification defenses in preprocessor-blind and white-box scenarios, and can do so while fully preserving benign accuracy in the former. Moreover, our results show that not only is FlowPure a highly effective purifier but it also holds strong potential for adversarial detection, identifying preprocessor-blind PGD samples with near-perfect accuracy. Our code is publicly available at \url{https://github.com/DistriNet/FlowPure}.
\end{abstract}

\section{Introduction}
\label{intro}

Deep neural networks continue to achieve remarkable success in various domains, yet remain vulnerable to adversarial attacks~\cite{DBLP:journals/corr/SzegedyZSBEGF13,DBLP:journals/corr/GoodfellowSS14}: crafted perturbations to inputs that lead to incorrect model predictions. Recently, adversarial purification (AP)~\cite{pixeldefend,diffpure,yoon2021adversarialpurificationscorebasedgenerative} has gained increasing attention as a promising strategy to enhance adversarial robustness. In contrast to Adversarial Training (AT) \cite{madry2018towards} techniques, which demand retraining a classifier and often come at the cost of benign performance, purification approaches aim to remove adversarial perturbations during inference. The remarkable success of diffusion models~\cite{ho2020denoisingdiffusionprobabilisticmodels} as generative models has placed them at the forefront of adversarial purification~\cite{yoon2021adversarialpurificationscorebasedgenerative,diffpure,li2025adbmadversarialdiffusionbridge}. 


The core goal of diffusion models~\cite{ho2020denoisingdiffusionprobabilisticmodels} is to transform a pure Gaussian into a complex data distribution  through two processes: (i) a forward process that iteratively adds noise to a sample until it becomes pure noise and (ii) a reverse process that gradually denoises samples toward the original data distribution.
In the context of adversarial purification, the forward process serves to dilute adversarial perturbations by injecting noise, whereas the reverse process reconstructs clean samples by removing the combined effect of injected Gaussian noise and adversarial noise.

However, recent advances in generative modeling have explored alternatives to diffusion models for learning transformations over complex data distributions. Flow Matching (FM)~\cite{lipman2023flow,tong_conditional_2023}, a modern framework for training Continuous Normalizing Flows (CNFs)~\cite{CNF1}, learns a continuous transformation between arbitrary source and target distributions by directly modeling the underlying dynamics. Unlike diffusion models, which rely on a stochastic process that gradually corrupts data with noise before learning a denoising procedure, Flow Matching learns a deterministic mapping between general distributions without requiring explicit noise injection. 

\begin{figure*}
    \centering   \includegraphics[width=1\linewidth]{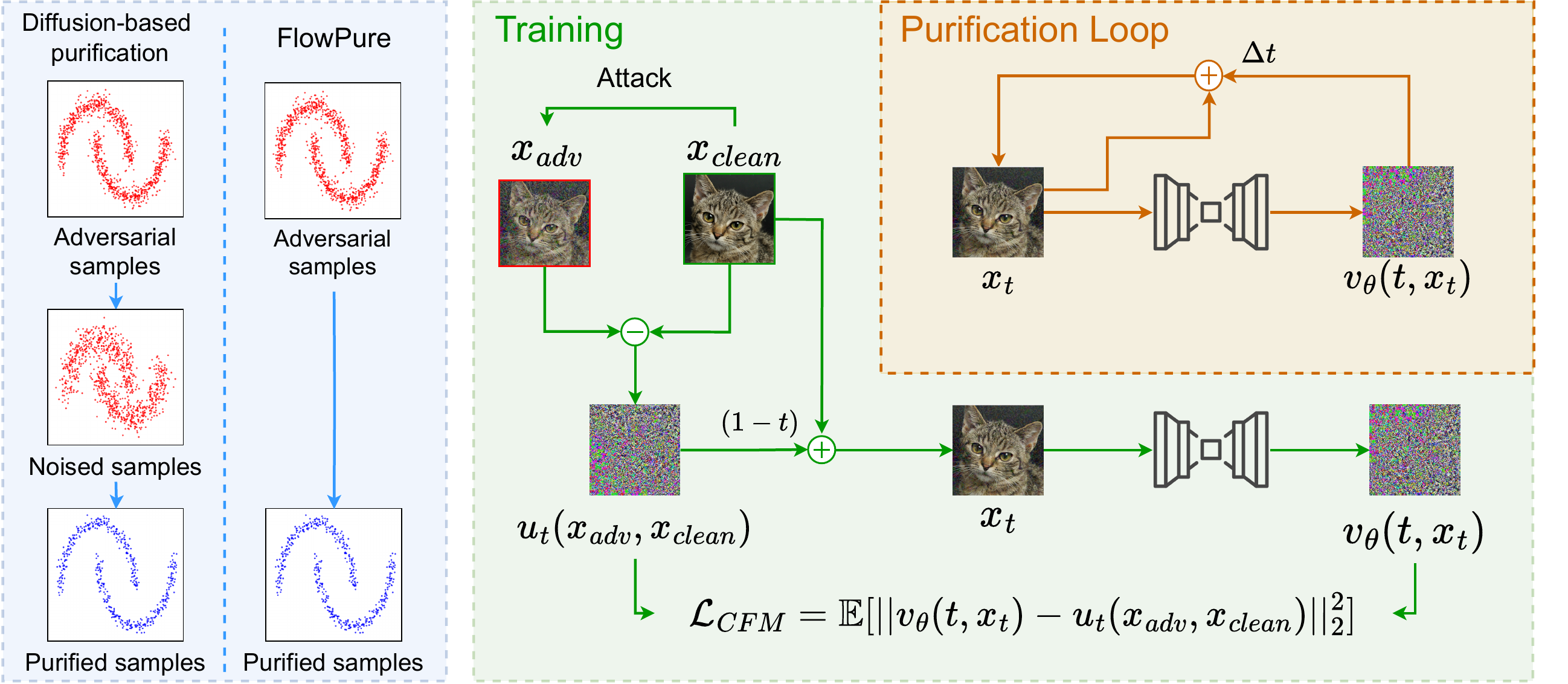}
    \caption{\textbf{Left:} Opposed to traditional diffusion-based purification methods that require adding Gaussian noise to samples, FlowPure can learn a direct transformation between arbitrary distributions. \textbf{Right:} Training and purification pipelines of FlowPure. The model $v_\theta(t,x_t)$ is trained to approximate the velocity field $u_t(x_{adv},x_{clean})$ using the loss $\mathcal{L}_{CFM}$. The purification loop illustrates the update rule, which should be applied iteratively, moving the sample toward the clean distribution.}

    \label{fig:training-diagram}
\end{figure*}

In this work, we introduce \textit{FlowPure}~\footnote{Our code is available at https://github.com/distrinet/flowpure.}, a novel adversarial purification method that leverages Flow Matching to directly transform adversarial examples into their clean counterparts (see Fig.~\ref{fig:training-diagram}). To the best of our knowledge, FlowPure is the first method to apply CNFs to adversarial purification. Whereas prior purification-based techniques lack mechanisms to exploit attack knowledge, FlowPure can \emph{incorporate it by sampling adversarial examples during training}. Although this may limit generalizability, our results indicate that it significantly improves performance under known threat models. To address scenarios involving unknown or adaptive attacks, we introduce \emph{a stochastic variant of FlowPure}. This attack-agnostic variant outperforms existing state-of-the-art diffusion-based purification methods, demonstrating that FlowPure can remain effective even without explicit access to the attack distribution. 


We consider two distinct adversarial threat models, with a primary focus on a preprocessor-blind setting. In this scenario, the attacker has full white-box access to the victim classifier, but is unaware of the existence of the purification mechanism, serving effectively as an upper bound on attack capabilities on black-box preprocessor-blind settings. Given the increasing prevalence of open-weight models in real-world deployments, we deem this threat model to be especially important, as it closely aligns with how attacks would typically be executed in practice, where interaction is often minimal and limited to model probing~\cite{grosse2024towards}. 

To better understand the limitations of adversarial purification, we also evaluate FlowPure in a fully adaptive white-box setting, where the attacker is assumed to have full gradient access to both the classifier and the defense mechanism. This represents a worst-case scenario, and in practice, it renders most defenses ineffective~\cite{tramer2020adaptive}. Recent work, such as DiffHammer~\cite{wang2024diffhammer}, has demonstrated that the robustness of diffusion-based purification methods has been significantly overestimated. In line with these findings, we observe that all purification-based methods, including FlowPure, struggle in this setting. Nevertheless, our stochastic variant shows improved performance under DiffHammer compared to existing approaches, outperforming state-of-the-art purifiers, even when all methods experience a sharp decrease in robustness.

Our main contributions are summarized as follows:

\begin{itemize}[leftmargin=1.12em]
    \item We introduce \textit{FlowPure}, a novel adversarial purification method that leverages Continuous Normalizing Flows to purify adversarial examples. FlowPure can directly map between arbitrary distributions, enabling the use of task-specific perturbations during training.

    \item We propose two variants of FlowPure: 1) a \textit{deterministic} variant that can rely on existing knowledge of attacks for purification of known threats, and 2) a \textit{stochastic} variant that relies on Gaussian noise for broad-spectrum robustness.

    \item We evaluate FlowPure across several settings and attack strategies, and show that our deterministic variant achieves state-of-the-art performance in preprocessor-blind scenarios with no degradation in standard accuracy, and that our stochastic variant outperforms existing purification methods in worst-case fully adaptive white-box attacks.
    
    \item Finally, we demonstrate that FlowPure can potentially serve as an effective adversarial detector, achieving near-perfect detection accuracy against oblivious PGD attacks, even at low perturbation levels. 
    
\end{itemize}

\section{Preliminaries \& Related Work}

This section provides background on diffusion models and prior state-of-the-art diffusion purification techniques. We then introduce Continuous Normalizing Flows, which form the foundation of our proposed method. Lastly, we briefly review relevant work on adversarial detection.

\subsection{Diffusion Purification}

Adversarial purification aims to remove perturbations introduced by an attacker, often by leveraging generative models to project adversarial examples onto the learned clean data distribution. Recently, diffusion models have emerged as a powerful alternative for adversarial purification. Diffusion models~\cite{ho2020denoisingdiffusionprobabilisticmodels} learn a transformation from a Gaussian to a complex data distribution. Training is performed by perturbing data with increasing noise levels and learning to predict the added noise at each time step. At inference time, new samples are generated by gradually removing predicted noise starting from Gaussian noise. We categorize purification algorithms leveraging diffusion into several types, relative to their working principles:



\textbf{Denoising purification.} DiffPure~\cite{diffpure} and ADP~\cite{yoon2021adversarialpurificationscorebasedgenerative} inject noise into adversarial samples and then use a diffusion model to remove it. The rationale behind these approaches is that, as more noise is introduced to samples, the distribution of diffused adversarial examples increasingly resembles that of diffused clean data and, as such, denoising is likely to recover a clean data sample. However, noise injection often disrupts semantically meaningful information, resulting in a drop in standard accuracy.

\textbf{Guided purification.} With the objective of addressing the limitations of standard denoising, GDMP~\cite{wang2022guided} incorporates guidance, ensuring that purified samples remain close to the original input, while still mitigating adversarial perturbations.

\textbf{Likelihood maximization.} Likelihood maximization techniques, such as ScoreOpt~\cite{zhang2023scoreopt} and LM~\cite{chen2024RDC}, leverage the fact that diffusion models approximate the score of the data distribution. These methods purify adversarial examples by maximizing their likelihood under the learned data distribution.

\textbf{Adversarial fine-tuning.} By identifying sub-optimal design choices in the design of traditional denoising purification methods, ADBM~\cite{li2025adbmadversarialdiffusionbridge} aims to enhance the white-box robustness by incorporating adversarial noise during the training of the diffusion model. This is achieved by fine-tuning a standard diffusion model with adversarial noise and then applying DiffPure~\cite{diffpure} with their fine-tuned model during inference. While both ADBM and FlowPure can incorporate adversarial noise during training, the former remains a diffusion model relying on a Gaussian forward process, while FlowPure \emph{can directly map arbitrary adversarial distributions}.

Even despite initial promise, the robustness of diffusion-based purification has been shown to be frequently overestimated~\cite{lee2023robustevaluationdiffusionbasedadversarial,kang2024diffattack,wang2024diffhammer,lucas2023randomness}. In white-box settings, diffusion-based purification relies on deep stochastic computation graphs, which can lead to gradient obfuscation and unreliable robustness estimates. While adaptive attacks such as BPDA~\cite{athalye2018obfuscatedgradientsfalsesense} and surrogate methods~\cite{lee2023robustevaluationdiffusionbasedadversarial} mitigate these issues, recent work demonstrates that more carefully designed attacks are more effective. In particular, DiffHammer~\cite{wang2024diffhammer}, the current state-of-the-art adaptive attack for diffusion-based purification, identifies and exploits vulnerable purification trajectories, outperforming prior evaluation methods.

\subsection{Flow Matching}
\label{sec:cnf}

Continuous Normalizing Flows (CNF)~\cite{CNF1} are a class of generative model that learns a transformation from a source distribution to a target distribution. The model is defined by a time-dependent velocity field $v_\theta(x,t)$, which induces a continuous transformation through an ODE. Traditional CNFs require simulating the ODE during training, resulting in high computational costs and poor scalability~\cite{CNF1}. This limitation is overcome by Flow Matching (FM)~\cite{liu2023flow,lipman2023flow}  by reformulating training as a regression problem. Instead of simulating the ODE, the model directly learns the velocity field that transports samples along a predefined probability path between source and target distributions. Conditional Flow Matching (CFM)~\cite{tong_conditional_2023} further simplifies this by conditioning on paired samples 
$(x_0, x_1)$ from source and target, which allows training from arbitrary source distributions 
without requiring access to intermediate marginal distributions. Importantly, the choice of 
conditional path constitutes a degree of freedom; in this work we adopt Optimal Transport (OT) 
paths~\cite{liu2023flow,tong_conditional_2023}, which correspond to straight-line trajectories 
and yield more stable and efficient inference compared to curved diffusion paths. Under 
this choice, the CFM objective becomes:
\begin{equation}
    \mathcal{L}_{\mathrm{CFM}}(\theta) = \mathbb{E}_{t,\, (x_0, x_1)} 
    \left\| v_\theta(x_t, t) - (x_1 - x_0) \right\|^2,
    \label{eq:cfm-loss}
\end{equation}
where $x_t = (1-t)x_0 + tx_1$ is a linear interpolation between source and target sample and $t \sim \mathcal{U}[0,1]$.

\subsection{Detection}

While adversarial purification methods aim to recover clean inputs from perturbed ones, an orthogonal line of defense focuses on detecting adversarial examples before they are processed by the classifier. The goal of adversarial detection is to identify inputs that have been maliciously crafted, enabling a system to either reject or flag them for further inspection. Most detection methods leverage intrinsic properties of the inputs that distinguish adversarial examples from clean data~\cite{detect1,detect2,detect3,detect4,beyond}.

Among these, BEYOND~\cite{beyond} introduces a self-supervised approach that compares an input to its augmented neighbors based on two criteria:  \textit{label consistency} and  \textit{representation similarity}. Clean samples tend to maintain consistent predictions and exhibit high feature similarity with their neighbors, whereas adversarial examples exhibit bigger deviations. 

\section{Methodology: FlowPure}
\label{sec:methodology}

In this section, we present FlowPure, our approach leveraging Continuous Normalizing Flows (CNF) for adversarial purification. We consider two variants: a deterministic variant that explicitly learns a mapping from adversarial to clean samples, and a Gaussian variant that leverages a pretrained generative CNF model together with a novel stochastic inference procedure. Finally, we discuss how FlowPure can also be used for adversarial detection.

\subsection{Deterministic FlowPure}
\label{sec:det-flowpure}

\begin{algorithm}[tb!]
\caption{Stochastic FlowPure}
\label{alg:stoch-flowpure}
\begin{algorithmic}[1]
\Require adversarial input $\tilde{x} \in [0,1]^d$, pretrained CNF model $v_\theta$, 
         starting timestep $t^*$, number of steps $N$, noise parameter $\alpha$

\State $x_{t^*} \leftarrow t^* \cdot \tilde{x} + (1 - t^*) \cdot \varepsilon_0$, \quad $\varepsilon_0 \sim \mathcal{N}(0,I)$ \Comment{Noise toward Gaussian source}
\State $\{t_1, \ldots, t_{N+1}\} \leftarrow \texttt{linspace}(t^*, 1, N+1)$ \Comment{Integration schedule}

\For{$i = 1, \ldots, N$}
    \State $t \leftarrow t_i$, \quad $t' \leftarrow t_{i+1}$
    \State $\eta \leftarrow (\alpha t + 1 - \alpha) / t$ \Comment{Time-dependent noise parameter}
    \State $t_\eta \leftarrow \alpha t + 1 - \alpha$ \Comment{Noised-back timestep}
    \State $\sigma \leftarrow \sqrt{(1 - \eta t)^2 - \eta^2(1-t)^2}$
    \State $x_{t_\eta} \leftarrow \eta \cdot x_t + \sigma \cdot \varepsilon_i$, \quad $\varepsilon_i \sim \mathcal{N}(0,I)$ \Comment{Inject noise}
    \State $x_{t'} \leftarrow x_{t_\eta} + (t' - t_\eta) \cdot v_\theta(x_{t_\eta}, t_\eta)$ \Comment{Euler step forward}
\EndFor

\State \Return $\hat{x}$
\end{algorithmic}
\end{algorithm}

The deterministic variant of FlowPure learns a direct continuous transformation from adversarial 
examples to clean samples by framing purification as a distribution transport problem. 
Training pairs $(x_0, x_1)$ 
are constructed by applying adversarial attacks to clean images. The model is then trained with 
$\mathcal{L}_{\mathrm{CFM}}$ (Eq.~\ref{eq:cfm-loss}) to regress the velocity field that 
transports adversarial samples to their clean counterparts along Optimal Transport (OT) paths.

To encourage robustness across threat levels rather than overfitting to a single attack 
configuration, we randomize attack parameters during training. Most importantly, we vary the 
perturbation budget $\epsilon \sim \mathcal{U}[0, \epsilon_{\max}]$, which exposes the model to 
the full spectrum from near-clean samples to strongly perturbed ones and implicitly encourages 
the model to behave as an approximate identity mapping near the clean data manifold.

At inference, purification is performed by integrating the learned ODE forward from $t=0$ to 
$t=1$, with the potentially adversarial input $\tilde{x}$ as the initial condition, returning $\hat{x} = 
x_{t=1}$ as the purified sample. Since the inference process is fully deterministic, we refer to this variant as Deterministic FlowPure.

\subsection{Stochastic FlowPure}
\label{sec:stoch-flowpure}

The stochastic variant of FlowPure is conceptually closer to DiffPure~\cite{diffpure}. Rather 
than learning a direct mapping from a specific adversarial distribution, it assumes access to a pretrained generative CNF model that represents a transformation from a Gaussian distribution to the data distribution, trained using OT paths. This makes the variant attack-agnostic, at the cost of semantic information by the noising step.

Purification is performed by first pushing the adversarial input toward the Gaussian distribution 
by integrating the ODE backward in time from $t=1$ to some $t^* \in (0,1)$, i.e., computing 
$x_{t^*} = t^* \cdot x_1 + (1-t^*) \cdot x_0$, and then integrating forward from $t=t^*$ to 
$t=1$ to obtain the purified sample. However, CNFs follow deterministic trajectories, which can make them more susceptible to adaptive attacks. To mitigate this, we introduce a novel stochastic inference procedure (see Alg.~\ref{alg:stoch-flowpure}) that injects noise into intermediate states of the ODE solver.

Concretely, given an intermediate sample $x_t$ at time $t$, we noise it toward an earlier 
timestep by:
\begin{equation}
    \tilde{x}_{\eta t} = \eta \cdot x_t + \sigma \cdot \varepsilon, \quad \varepsilon \sim 
    \mathcal{N}(0, I),
    \label{eq:noise-step}
\end{equation}
where $\sigma = \sqrt{(1 - \eta t)^2 - \eta^2(1-t)^2}$ and $\eta \in (0, 1)$ controls how far back in time the sample is pushed. The simple derivation of $\sigma$ can be found in Section~\ref{app:baselines-conf} of the Appendix. To ensure 
that noise injection is proportional to the amount of noise already present in the intermediate 
sample, we make 
$\eta$ time-dependent:
\begin{equation}
    \eta(t) = \max\left(\frac{\alpha t + 1 - \alpha}{t},\ 0\right),
    \label{eq:eta}
\end{equation}
where $\alpha \geq 1$ is a tunable constant. This schedule for $\eta(t)$ ensures that the additional noise is linearly dependent on the amount of noise already present in the sample. The resulting stochastic purification paths increase 
uncertainty from the attacker's perspective, improving robustness against adaptive attacks.

\begin{figure*}[tb!]
    \centering
    \includegraphics[width=0.99\textwidth]{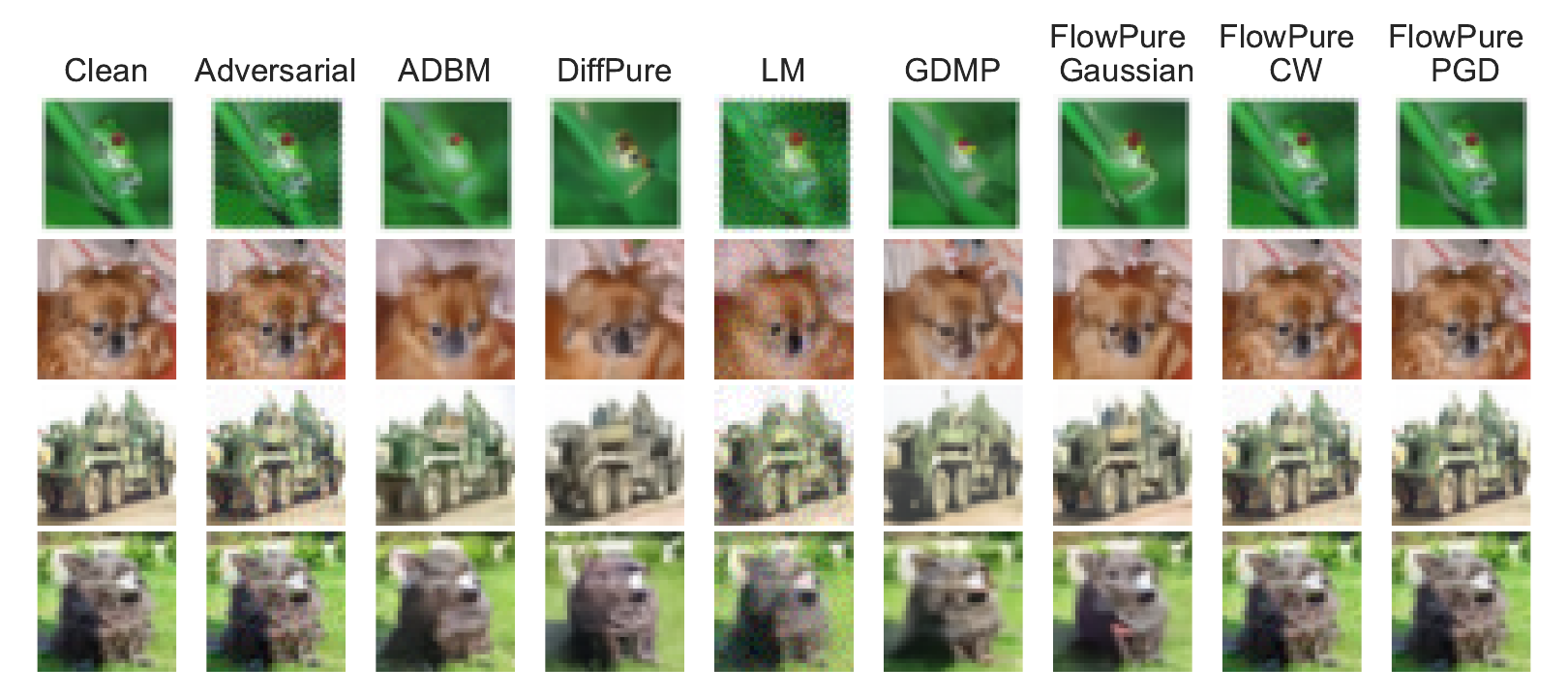}
    \caption{PGD adversarial examples in CIFAR-10 purified with several methods. $FlowPure^{PGD}$ produces the clearest reconstructions, often removing minor artifacts present in the original image. }
    \label{fig:visualization}
\end{figure*}

\subsection{Detection} 
We also explore the potential of FlowPure for adversarial detection. We leverage the magnitude of the learned velocity field $||{v}_\theta(t,x)||_2^2$ at time $t=0$, as a detection score. The underlying intuition is that adversarial examples, residing outside of the data manifold, exhibit a higher initial velocity compared to clean samples. This enables us to flag samples as adversarial if their initial velocity is above a threshold.

\section{Evaluation}
This section reports on a comprehensive evaluation of FlowPure under various threat models. We assess purification performance in both realistic (preprocessor-blind) and worst-case (fully adaptive white-box) settings. We also report preliminary results of adversarial detection with FlowPure.

\subsection{Experimental setup}
\label{sec:experimentalsetup}

\textbf{Baselines.}  We evaluate our approach on two benchmark datasets: CIFAR-10 and CIFAR-100~\cite{cifar}. The victim classifier (referred to as \emph{Victim} in our experiments) for both datasets and for each purifier is the widely used WideResNet-28-10~\cite{Zagoruyko2016WRN}. For comparison, we benchmark against multiple diffusion-based purification methods: DiffPure~\cite{diffpure}, GDMP~\cite{wang2022guided}, LM~\cite{chen2024RDC} and ADBM~\cite{li2025adbmadversarialdiffusionbridge}. While we do not consider AT methods, as they form a separate family of approaches focusing on hardening the model itself, these baselines cover the major diffusion-based purification paradigms explored in the literature and allow us to assess the relative performance of our method. All purification models, have the DDPM++~\cite{song2021scorebased} architecture, except our Gaussian variant that uses a standard U-Net~\cite{unet}, with the configuration of Lipman et al.~\cite{lipman2023flow}. More details are included in Section~\ref{app:compres} of the Appendix. To evaluate preprocessor-blind scenarios, we consider Projected Gradient Descent (PGD)~\cite{madry2018towards} and Carlini \& Wagner (CW)~\cite{cw2016attack} attacks, which operate under the $L_\infty$ and $L_2$ norms, respectively. A qualitative comparison of reconstruction-based adversarial defenses applied to PGD adversarial examples can be seen in Figure~\ref{fig:visualization}. To assess transferability of our method, we additionally evaluate on the individual attacks in the AutoAttack~\cite{croce2020reliable} benchmark, for both the $L_\infty$ and $L_2$ norms.  We also assess white-box robustness using BPDA~\cite{athalye2018obfuscatedgradientsfalsesense} and the state-of-the-art DiffHammer~\cite{wang2024diffhammer} attack. For detection, we report preliminary results in which we compare against BEYOND~\cite{beyond}. We consider BEYOND the most relevant baseline detector for our preliminary detection experiments, as it consistently outperforms prior methods. For both attacks and defenses, we use standard configurations as reported in the original works, unless specified otherwise (the details are outlined in Section~\ref{app:baselines-conf} of the Appendix). 

\textbf{Metrics.} We use two metrics to evaluate all purification approaches: \textit{standard accuracy} and \textit{robust accuracy}. Standard accuracy measures the accuracy of the defended model on clean unperturbed data. Robust accuracy measures the accuracy of the defended model on adversarial examples generated by adversarial attacks. For robust accuracy, we employ the framework of resubmission attacks described in DiffHammer~\cite{wang2024diffhammer} to study stochastic defenses. As such, we report \textit{average robust accuracy ($avg$)}, defined as the mean accuracy after $N$ resubmissions, and \textit{worst-case robust accuracy ($worst$)}, in which an attack is considered successful if the same adversarial example is misclassified once in $N$ resubmissions.  For all experiments, we consider $N=10$ as the number of resubmissions. Note that for deterministic defenses this has no impact. Moreover, and consistent with previous work, we evaluate the complete test set of each dataset for preprocessor-blind attacks, and a subset of $512$ images sampled at random for white-box attacks. Due to the small subset, we report the average and standard deviation of $3$ runs in the white-box setting. To evaluate detection, we use the area under the curve (AUC) of the receiver operating characteristic (ROC). 
Additionally we report the TPR at FPRs of $5\%$ , $3\%$ and $1\%$.

\textbf{FlowPure.} We train three CNF models for each dataset. The first, which we denote $FlowPure^{PGD}$, is trained on PGD samples. The second, $FlowPure^{CW}$, is trained on CW samples. The third, $FlowPure^{Gauss}$, corresponds to the Gaussian FlowPure variant.  In all cases, ODE integration is performed using Euler’s method. $\text{FlowPure}^{\text{Gauss}}$ 
additionally employs intermediate noise injection as described in Algorithm~\ref{alg:stoch-flowpure}, 
with $t^*=0.85$ and $\alpha=1.5$ unless otherwise specified. Further implementation details, including the distributions used for attack parameters, are provided in Section~\ref{app:baselines-conf} of the Appendix.


\subsection{Preprocessor-blind evaluation}
\label{sec:preprocesssor_blind_eval}

\begin{table*}[t!]
\centering
    \renewcommand{\arraystretch}{1.1}

    \caption{Purification on PGD and CW attacks in the preprocessor-blind setting, and under BPDA~\cite{athalye2018obfuscatedgradientsfalsesense} and DH~\cite{wang2024diffhammer} in the fully-adaptive white-box setting. Metrics report average accuracy ($avg$) and worst-case accuracy ($worst$) after 10 resubmissions (for $FlowPure^{PGD}, FlowPure^{CW}$ and \textit{Victim} the two metrics coincide, as they are deterministic). The top-performing method is shown in bold, the second-best is underlined. BPDA and DH do not apply to the victim in isolation.}
    \label{tab:main_results}

    \tiny
\begin{tabular}{@{}lllcclcccclllll@{}}
\toprule
 & \textbf{Method} &  & \multicolumn{2}{c}{\textbf{\begin{tabular}[c]{@{}c@{}}Standard\\  Accuracy\end{tabular}}} &  & \multicolumn{9}{c}{\textbf{Robust Accuracy}} \\ \cmidrule(lr){4-5} \cmidrule(l){7-15} 
 &  &  & \multicolumn{1}{l}{} & \multicolumn{1}{l}{} &  & \multicolumn{4}{c}{\textbf{Preprocessor-blind}} &  & \multicolumn{4}{c}{\textbf{White-box}} \\ \cmidrule(lr){7-10} \cmidrule(l){12-15} 
 &  &  & \multicolumn{1}{l}{} & \multicolumn{1}{l}{} &  & \multicolumn{2}{c}{PGD} & \multicolumn{2}{c}{CW} &  & \multicolumn{2}{c}{DH} & \multicolumn{2}{c}{BPDA} \\
 &  &  & $avg$ & $worst$ &  & $avg$ & $worst$ & $avg$ & $worst$ &  & \multicolumn{1}{c}{$avg$} & \multicolumn{1}{c}{$worst$} & \multicolumn{1}{c}{$avg$} & \multicolumn{1}{c}{$worst$} \\ \midrule
\multirow{9}{*}{\rotatebox[origin=c]{90}{\textbf{CIFAR-10}}} & \textit{Victim} &  & $95.81$ & $95.81$ &  & $0.04$ & $0.04$ & $0.00$ & $0.00$ &  & \multicolumn{1}{c}{--} & \multicolumn{1}{c}{--} & \multicolumn{1}{c}{--} & \multicolumn{1}{c}{--} \\
 & \textit{Baselines:} &  &  &  &  &  &  &  &  &  &  &  &  &  \\
 & \:\: DiffPure~\cite{diffpure} &  & $89.99$ & $73.69$ &  & $89.10$ & $72.25$ & $89.82$ & $73.76$ &  & $\underline{40.64 \pm 1.35}$ & \underline{$17.70 \pm 1.81$} & $71.74 \pm 0.51$ & $41.40 \pm 2.62$ \\
 & \:\: GDMP~\cite{wang2022guided} &  & $92.41$ & $80.81$ &  & $\underline{91.48}$ & $78.67$ & $92.37$ & $80.59$ &  & $33.78 \pm 1.02$ & $14.90 \pm 1.25$ & \underline{$73.23 \pm 0.62$} & \contour{black}{$47.46 \pm 2.30$} \\
 & \:\: LM~\cite{chen2024RDC} &  & $82.94$ & $65.03$ &  & $67.61$ & $44.86$ & $78.70$ & $56.20$ &  & $12.55 \pm 0.39$ & $4.82 \pm 0.23$ & $50.96 \pm 0.56$ & $24.86 \pm 0.98$ \\
 & \:\: ADBM~\cite{li2025adbmadversarialdiffusionbridge} &  & $91.93$ & $78.77$ &  & $89.08$ & $72.59$ & $90.25$ & $74.86$ &  & $32.35 \pm 1.42$ & $13.15 \pm 0.62$ & $65.25 \pm 0.76$ & $36.26 \pm 1.30$ \\
 & \textsl{Ours}: &  &  &  &  &  &  &  &  &  &  &  &  &  \\
 & \:\: FP$^{\text{PGD}}$ &  & $\mathbf{96.00}$ & $\mathbf{96.00}$ &  & $\mathbf{92.23}$ & $\mathbf{92.23}$ & $\underline{91.45}$ & $\underline{91.45}$ &  & $1.17 \pm 0.20$ & $1.17 \pm 0.20$ & $13.34 \pm 1.19$ & $13.34 \pm 1.20$ \\
 & \:\: FP$^{\text{CW}}$ &  & $\underline{95.96}$ & $\underline{95.96}$ &  & $87.33$ & $\underline{87.33}$ & $\mathbf{94.36}$ & $\mathbf{94.36}$ &  & $1.11 \pm 0.49$ & $1.11 \pm 0.49$ & $23.43 \pm 0.39$ & $23.43 \pm 0.40$ \\
 & \:\: $\text{FP}^{\text{Gauss}}$ &  & $89.60$ & $74.59$ & \multicolumn{1}{c}{} & $88.81$ & $72.79$ & $89.44$ & $74.31$ &  & \contour{black}{$43.12 \pm 1.49$} & \contour{black}{$21.87 \pm 1.28$} & \contour{black}{$73.95 \pm 0.65$} & \underline{$46.09 \pm 1.40$} \\ \midrule
\multirow{9}{*}{\rotatebox[origin=c]{90}{\textbf{CIFAR-100}}} & \textit{Victim} &  & $80.73$ & $80.73$ &  & $0.01$ & $0.01$ & $0.00$ & $0.00$ &  & \multicolumn{1}{c}{--} & \multicolumn{1}{c}{--} & \multicolumn{1}{c}{--} & \multicolumn{1}{c}{--} \\
 & \textit{Baselines:} &  &  &  &  &  &  &  &  &  &  &  &  &  \\
 & \:\: DiffPure~\cite{diffpure} &  & $61.65$ & $37.71$ &  & $\underline{60.03}$ & $35.23$ & $60.78$ & $36.90$ &  & \underline{$15.87 \pm 0.61$} & \underline{$5.20 \pm 1.07$} & $35.52 \pm 1.53$ & $14.12 \pm 0.88$ \\
 & \:\: GDMP~\cite{wang2022guided} &  & $66.26$ & $44.28$ &  & $\mathbf{64.39}$ & $\underline{42.16}$ & $65.46$ & $42.43$ &  & $11.77 \pm 0.95$ & $4.49 \pm 0.67$ & \contour{black}{$37.27 \pm 1.98$} & \contour{black}{$16.08 \pm 1.47$} \\
 & \:\: LM~\cite{chen2024RDC} &  & $45.56$ & $23.23$ &  & $26.89$ & $10.98$ & $39.98$ & $17.99$ &  & $3.35 \pm 0.98$ & $0.46 \pm 0.30$ & $13.05 \pm 0.28$ & $4.68 \pm 0.39$ \\
 & \:\: ADBM~\cite{li2025adbmadversarialdiffusionbridge} &  & $64.76$ & $39.07$ &  & ${57.43}$ & ${29.99}$ & $60.24$ & $32.72$ &  & $7.13 \pm 0.45$ & $1.49 \pm 0.22$ & $23.52 \pm 1.25$ & $6.51 \pm 0.96$ \\
 & \textit{Ours:} &  &  &  &  &  &  &  &  &  &  &  &  &  \\
 & \:\: FP$^{\text{PGD}}$ &  & $\underline{80.75}$ & $\underline{80.75}$ &  & ${52.90}$ & $\mathbf{52.90}$ & $\underline{66.67}$ & $\underline{66.67}$ &  & $0.38 \pm 0.20$ & $0.38 \pm 0.20$ & $2.67 \pm 0.92$ & $2.67 \pm 0.93$ \\
 & \:\: FP$^{\text{CW}}$ &  & $\mathbf{80.84}$ & $\mathbf{80.84}$ &  & $28.67$ & $28.67$ & $\mathbf{75.51}$ & $\mathbf{75.51}$ &  & $0.26 \pm 0.11$ & $0.26 \pm 0.11$ & $2.47 \pm 0.68$ & $2.47 \pm 0.69$ \\
 & \:\: $\text{FP}^{\text{Gauss}}$ &  & $60.75$ & $37.31$ & \multicolumn{1}{c}{} & $59.57$ & $35.76$ & $60.65$ & $37.07$ &  & \contour{black}{$16.06 \pm 0.29$} & \contour{black}{$5.33 \pm 0.92$} & \underline{$37.22 \pm 1.65$} & \underline{$15.10 \pm 0.56$} \\ \bottomrule
\end{tabular}
\end{table*}

We first evaluate our method in the preprocessor-blind setting, in order to evaluate how effectively FlowPure defends against known threats. Table~\ref{tab:main_results} displays the results for this scenario in CIFAR-10 and CIFAR-100. We also examine the extent to which the robustness of our deterministic variants transfers to attacks and norms that differ from their training distribution. Lastly, we provide a preliminary evaluation of the use of FlowPure as a detector of adversarial examples. 

 \textbf{Benign performance.} FlowPure (in its deterministic variants $FlowPure^{PGD}$ and $FlowPure^{CW}$), presents no degradation in the accuracy of clean samples, as opposed to baselines. Furthermore, there is even a slight improvement in the metric. We theorize that this could be due to FlowPure learning a more robust mapping of clean data, as a form of implicit regularization, or through mitigation of non-adversarial noise already present on unperturbed samples. This interpretation is supported by visual exploration of purified samples (see Fig.~\ref{fig:visualization}), where FlowPure reconstructions appear cleaner and more faithful to the original inputs, often removing spurious noise while preserving semantic content.

\textbf{Robustness.} FlowPure consistently exhibits superior robustness in all \emph{attack-aware scenarios} (when the evaluated attack matches the one used in training) but one (stochastic defenses achieve a higher average robust accuracy in CIFAR-100 for PGD, with significant degradation in benign performance), suggesting that the method can effectively reverse adversarial perturbations in a targeted manner. 
 
 \textbf{Transferability.} In \emph{transfer scenarios} (the attack does not match the one used during training), $FlowPure^{PGD}$ demonstrates strong transferability to CW attacks. $FlowPure^{CW}$, however, struggles to generalize to PGD, displaying a notable drop in robustness in CIFAR-100. This asymmetry likely arises from the structural differences between PGD and CW perturbations, as PGD perturbations tend to be larger in magnitude and more distributed, yielding a more generalizable mapping. In contrast, CW perturbations, while smaller in magnitude, are highly specialized, leading to increased complexity for $FlowPure^{CW}$ in effectively counteracting broader perturbations introduced by PGD attacks. 

\begin{figure*}[t]
    \centering
    \includegraphics[width=0.99\textwidth]{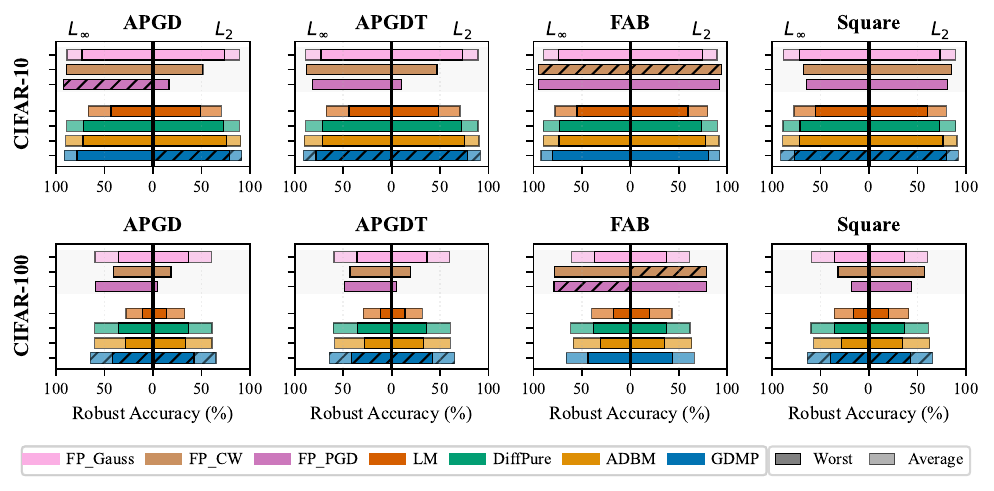}
    \caption{Robustness against individual attacks of the AutoAttack benchmark, in the preprocessor-blind setting. We include both the $L_\infty$ norm (left) and the $L_2$ norm (right). The top performing method is shown with a pattern. }
    \label{fig:AA_grid}
\end{figure*}

To further assess the transferability of FlowPure to unseen attacks and norms, we evaluate its performance against the individual attacks of the AutoAttack~\cite{croce2020reliable} benchmark suite in the preprocessor-blind setting. Results are reported in Figure~\ref{fig:AA_grid}. We refer to Section~\ref{app:addexpres} of the Appendix for the complete results.

We observe that Gaussian-based methods are generally more consistent in terms of transferability compared to deterministic variants, as the latter are tailored to specific attacks.  Among deterministic models, $FlowPure^{PGD}$ shows some transferability when facing attacks of a similar type within the same norm, but cross-norm transferability is more limited, with the exception of strong performance on cross-norm FAB. By contrast, $FlowPure^{CW}$ demonstrates stronger cross-norm transferability but weaker performance against unseen attacks (with the exception of FAB), reinforcing trends observed in earlier experiments. While further evaluation is needed, these results suggest that deterministic variants may offer some degree of transferability to unseen attacks, but without strong guarantees.

 \textbf{Detection.} We additionally explore the use of FlowPure for detection of adversarial examples in preprocessor-blind non-transfer scenarios. As a baseline, we compare FlowPure to BEYOND~\cite{beyond}, a recent state-of-the-art detection method based on Self-Supervised Learning (SSL). {BEYOND detects adversarial examples by examining the label consistency and representation similarity between an input and its augmented neighbors. The results are summarized in Figure \ref{fig:detection_comparison}. For both PGD and CW attacks, we evaluate the area under the ROC curve (AUC), and the true positive rate (TPR) at several false positive rate (FPR) thresholds.  
 
When considering PGD attacks, our results consistently show that FlowPure outperforms BEYOND on both CIFAR-10 and CIFAR-100, with FlowPure achieving near-perfect AUC scores and substantially higher TPR at low FPRs across perturbation levels. Moreover, FlowPure becomes increasingly effective in distinguishing more strongly perturbed PGD samples, whereas BEYOND's detection capabilities appear to plateau or even slightly decrease in relative terms under larger perturbations. Conversely, when dealing with the smaller perturbations of CW attacks, BEYOND is generally more effective, except at very small FPRs on CIFAR-100, due to its use of augmentations. We defer additional results, including transfer scenarios, to Section~\ref{app:addexpres} of the Appendix.

\begin{figure*}[]
    \centering
    \begin{subfigure}[t]{0.49\linewidth}
        \vspace{0pt}
        \includegraphics[width=\textwidth]{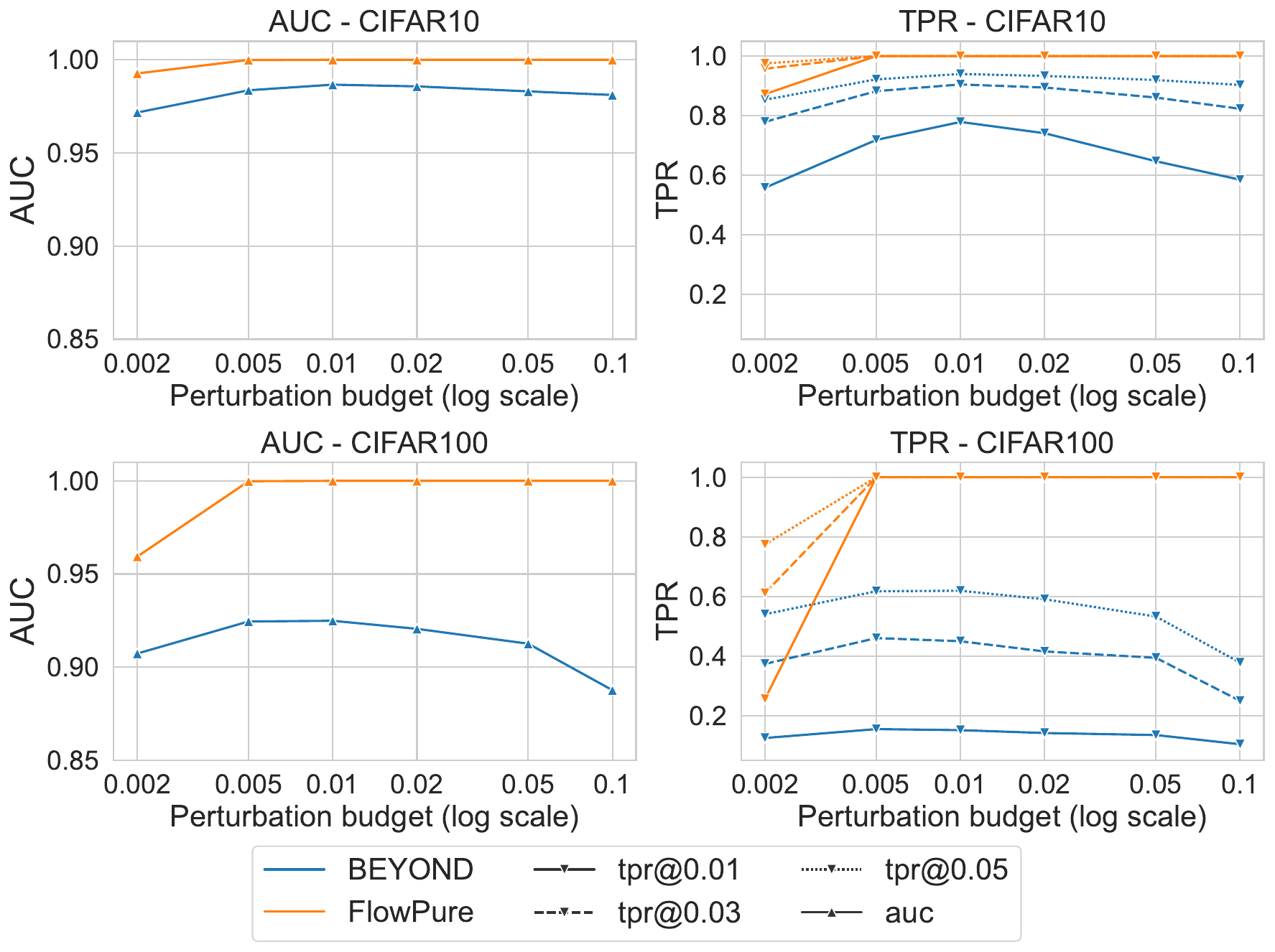}
        \caption{Detection of PGD attacks. AUC and TPR@FPR are shown for different perturbation levels. }
    \end{subfigure}
    \hfill 
    \begin{subfigure}[t]{0.49\linewidth}

        \vspace{0pt}
        \includegraphics[width=\textwidth]{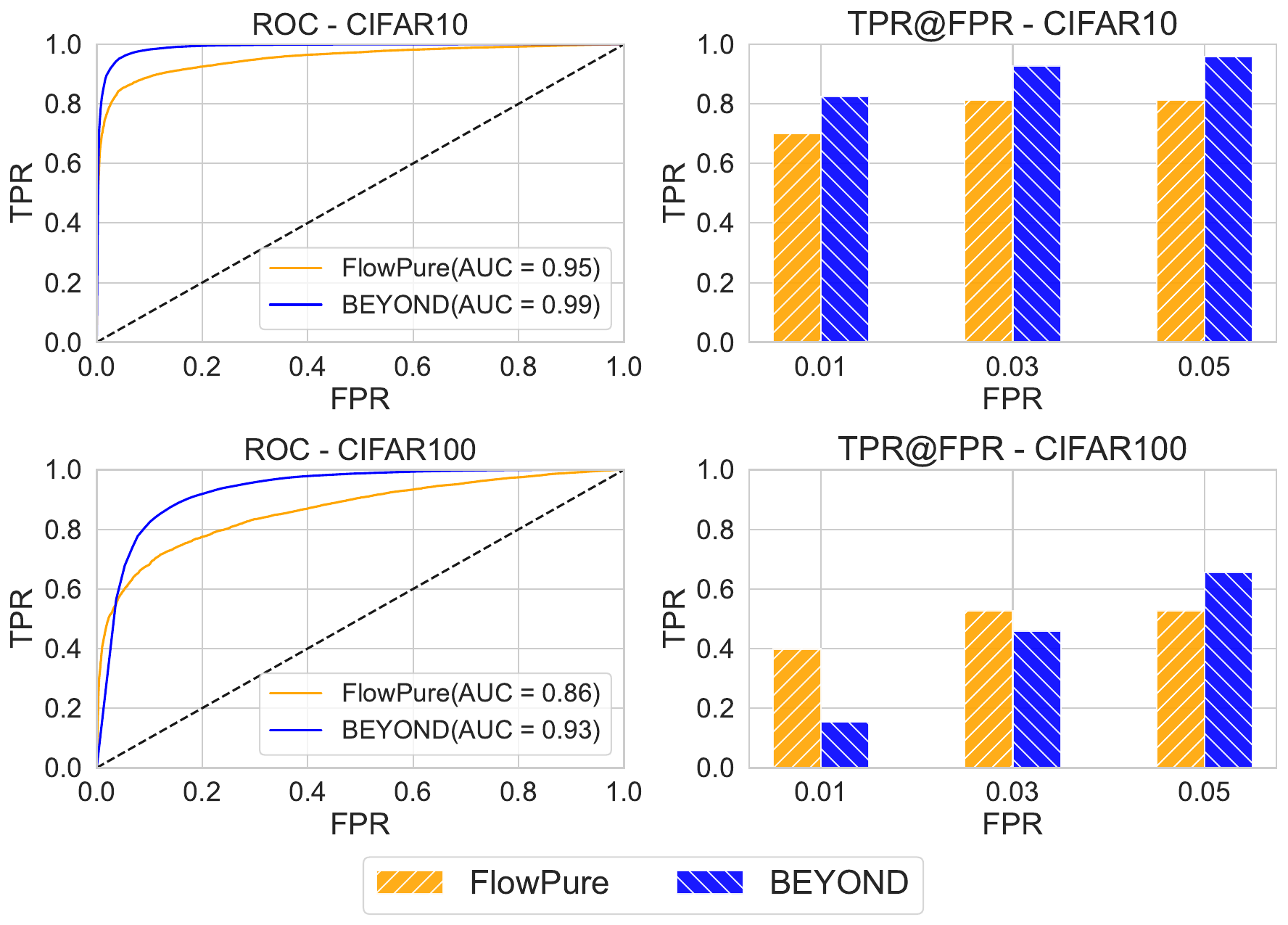}
       \captionsetup{skip=8pt}

        \caption{Detection of CW attacks. ROC curve and TPR@FPR are shown. }
    \end{subfigure}

    \caption{Detection results of FlowPure and BEYOND in the preprocessor-blind setting.}
    \label{fig:detection_comparison}
\end{figure*}

\subsection{Fully adaptive white-box evaluation}
\label{sec:white_box_eval}

We further evaluate our approach in a fully adaptive white-box setting, in order to determine how well FlowPure performs in worst-case conditions. In this scenario, the attacker is assumed to have gradient access to both the classifier and the purification mechanism. As attacks, we rely on BPDA~\cite{athalye2018obfuscatedgradientsfalsesense} and DiffHammer~\cite{wang2024diffhammer} to provide an upper bound for the robustness of the entire system.

The results, also detailed in Table~\ref{tab:main_results}, reveal a substantial degradation in performance compared to the preprocessor-blind scenario, given all methods suffer significant decreases in robust accuracy. This trend is most apparent in the case of our deterministic variants, as the deterministic mapping through CNFs may be more susceptible to gradient-based exploitation than stochastic methods which introduce noise that can disrupt precise adversarial gradients by design.

Importantly, BPDA represents a weaker adaptive attack because it uses approximate gradients, as opposed to DH, which uses exact gradients of the surrogate purification process. Under BPDA, we observe that only GDMP outperforms our method on CIFAR-100. Furthermore the gap between BPDA and DH is smaller for FlowPure, indicating that there is less gradient masking in our method. Under the stronger attack (DH), our Gaussian-trained variant consistently achieves the best performance across both datasets.

Although $FlowPure^{Gauss}$ exhibits the strongest performance under fully adaptive white-box attacks, it is important to note that none of the evaluated methods demonstrate practical viability in this setting. With robust accuracy metrics falling to 43\% in CIFAR-10 and below 17\% on CIFAR-100, across all approaches, no defense proves effective against adversaries with full gradient access, reinforcing the broader challenge of securing models under such extreme threat conditions.

\textbf{Ablation study.} To better understand the performance of Gaussian FlowPure, we also conduct a comprehensive analysis evaluating its performance against DH~\cite{wang2024diffhammer} across varying levels of noise injection. Moreover, we benchmark it against DiffPure, the most comparable diffusion-based purification method. This setup allows us to isolate the impact of using CNFs for purification. Both FlowPure and DiffPure use $10$ iterations to remove injected noise, and the attacker uses a surrogate process with $5$ iterations. Our results, presented in Figure \ref{fig:ablation}, show that Gaussian FlowPure consistently achieves higher robust accuracy than DiffPure across a range of noise levels, while incurring similar or lower degradation in benign accuracy. The complete results, including our preprocessor-blind scenario, are available in Section~\ref{app:addexpres} of the Appendix.

\begin{figure*}[tb!]
    \centering
    \includegraphics[width=1\linewidth]{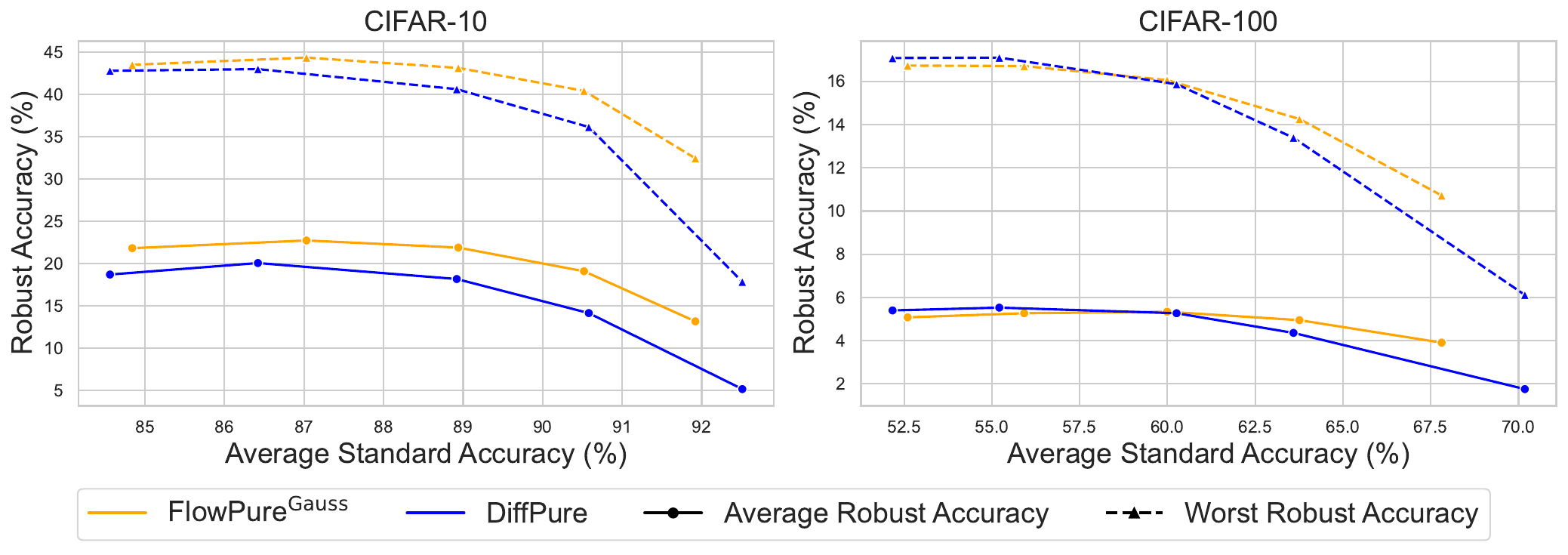}
    \caption{Standard/Robust accuracy trade-off under DH~\cite{wang2024diffhammer} (10 resubmissions), of $FlowPure^{Gauss}$ and DiffPure~\cite{diffpure}, for different levels of noise injection (ranges are $t^* \in [0.8,0.9]$ for the former, and $t^*\in [50,180]$ for the latter). Higher values on both axes are better, with increases in standard accuracy correlated with lower noise levels. Note that axes differ between the plots.}
    \label{fig:ablation}
\end{figure*}

\section{Discussion and Limitations}
\label{sec:discussion}

Our experimental results in the preprocessor-blind setting demonstrated the effectiveness of FlowPure against strong $L_\infty$ (PGD) and $L_2$ (CW) attacks. Moreover, the deterministic variants of FlowPure showed no degradation in benign performance, while providing significant robustness against PGD and CW attacks, highlighting their potential as a practical defense mechanism in scenarios where attackers are unaware of the purification process. Although incorporating attack-specific knowledge during training leads to strong empirical performance, this reliance can limit applicability when threat models are unknown or evolve over time, as our evaluations in the fully adaptive white-box setting demonstrated. Despite this, our experiments in transfer scenarios indicated encouraging generalization properties, suggesting that these methods can provide robustness beyond their specific training distribution.

Our evaluation under fully adaptive white-box attacks shows that robust accuracy drops sharply for all studied defenses, specially under DiffHammer~\cite{wang2024diffhammer}, reaffirming the challenge of defending against adversaries with full gradient access. This degradation is most pronounced in the case of our deterministic variants, which are more easily exploited with gradient-based attacks. In contrast, incorporating stochasticity during training and inference can indeed improve robustness in this setting, albeit at the cost of reduced benign accuracy, highlighting the inherent trade-off in stochastic preprocessing defenses~\cite{gao2022limitations}. Among all evaluated purifiers, Gaussian FlowPure achieves the strongest performance in worst-case white-box scenarios.
To further verify the benefits of flow matching-based purification, an evaluation of FlowPure on higher-complexity datasets (higher resolution, scale, and number of classes) is recommended. Crucially, such experiments are currently limited by computational constraints, given the lack of publicly available pretrained CNF models comparable to those widely available for diffusion.}


While white-box evaluations are crucial for understanding the theoretical limits of robustness, their practical utility as the primary benchmark for adversarial defense warrants reconsideration. Under fully adaptive white-box attacks, all studied purification-based defenses, including FlowPure, experience a consistent collapse. The substantial degradation in robustness in this setting indicates that current purification approaches do not offer meaningful protection against adversaries with full gradient access, a scenario where adversarial training remains comparatively more effective, albeit with well-known drawbacks, such as reduced benign accuracy and limited flexibility.

With this work, we underline the importance of redirecting efforts toward more realistic threat models, such as preprocessor-blind settings, that better reflect constraints faced by real-world adversaries. Regarding the evaluation of the entire purification pipeline, black-box attack scenarios where adversaries interact with the system exclusively through queries can also be considered. While such threat models may better reflect deployment environments, they often presuppose substantial query budgets, which may be prohibitive in practice. Nonetheless, incorporating black-box evaluations remains a meaningful avenue for future work.


\section{Conclusion}
\label{sec:conclusion}

In this work, we introduced FlowPure, a novel adversarial purification method leveraging Continuous Normalizing Flows trained with Conditional Flow Matching. FlowPure improves on state-of-the-art in multiple scenarios: for preprocessor-blind attacks, our approach directly learns the mapping between adversarial and clean data distributions, effectively removing adversarial perturbations. This key difference allows FlowPure to fully preserve benign accuracy on CIFAR-10 and CIFAR-100 datasets during preprocessor-blind attacks. For adaptive white-box attacks we found that our Gaussian variant of FlowPure outperforms state-of-the-art during purification under the strongest available attack.
Overall, FlowPure offers a realistic and deployable purification approach where benign performance is kept high while providing strong robustness.
Finally, we showcased the dual capability of FlowPure as a highly effective adversarial detector, as it achieves near-perfect detection accuracy against oblivious PGD attacks.

\section*{Acknowledgments}

This research is partially funded by the Internal Funds KU Leuven, and by the Cybersecurity Research Program Flanders. The authors thank Ilia Shumailov for providing valuable feedback on an earlier version of the manuscript.

%
%

\bibliographystyle{splncs04}
\bibliography{bib}

\newpage

\appendix

The appendix includes additional information such as preliminaries and configurations for both defenses and attacks (Section~\ref{app:baselines-conf}); additional experimental results on various noise levels, additional attacks and norms, and detection distribution plots (Section~\ref{app:addexpres}); and the utilized computational resources and technical implementation details for our experiments (Section~\ref{app:compres}).

\section{Preliminaries and Configurations}
\label{app:baselines-conf}

\subsection{Baseline defenses}

The baseline defense configurations used in our experiments largely follow those proposed in the original papers. We used implementations for each baseline (except BEYOND) provided by the authors of DiffHammer~\cite{wang2024diffhammer}\footnote{\url{https://github.com/Ka1b0/DiffHammer} (License: MIT)}.




\paragraph{DiffPure~\cite{diffpure}}
DiffPure purifies adversarial examples by first corrupting them with Gaussian noise through 
the diffusion forward process up to a predefined timestep $t^*$:
\begin{equation}
x_{t^*} = \sqrt{\bar{\alpha}_{t^*}} x_0 + \sqrt{1 - \bar{\alpha}_{t^*}} \epsilon,
\end{equation}
where $\alpha_t$ follows a predefined schedule, $\bar{\alpha}_t = \prod_{i=1}^t \alpha_i$, 
and $\epsilon \sim \mathcal{N}(0, I)$. The rationale is that sufficient noise injection 
causes the adversarial and clean distributions to converge, after which the standard DDPM 
reverse process recovers a clean sample:
\begin{equation}
x_{t-1} = \frac{1}{\sqrt{\alpha_t}} \left[ x_t - \frac{1 - \alpha_t}{1 - \bar{\alpha}_t} 
\epsilon_\theta(x_t, t) \right] + \sigma_t z,
\end{equation}
where $\epsilon_\theta(x_t, t)$ is a parameterized denoising network, $z \sim \mathcal{N}(0,I)$, 
and $\sigma_t = \frac{1 - \bar{\alpha}_{t-1}}{1 - \alpha_t}(1 - \bar{\alpha}_t)$.
We use $t^* = 100$. Following~\cite{lee2023robustevaluationdiffusionbasedadversarial}, we 
adopt a surrogate process with differing time intervals for adaptive attacks, set to $20$ 
for the attacker and $10$ for the defender.



\paragraph{GDMP~\cite{wang2022guided}}
GDMP augments the DiffPure reverse process with a similarity guidance term that encourages 
purified samples to remain close to the original input, trading off noise removal against 
semantic preservation:
\begin{equation}
\footnotesize
x_{t-1} = \frac{1}{\sqrt{\alpha_t}} \left[ x_t - \frac{1 - \alpha_t}{1 - \bar{\alpha}_t} 
\left( \epsilon_\theta(x_t, t) + s\sigma_t \nabla_{x_t} D(x_t, x_{t'}) \right) \right] 
+ \sigma_t z,
\end{equation}
where $s$ is a guidance scale and $x_{t'} = \sqrt{\bar{\alpha}_{t'}} x_0 + 
\sqrt{1 - \bar{\alpha}_{t'}} \epsilon$. We use $t^* = 36$ with $4$ successive purification 
rounds. Time intervals are set to $12$ for the attacker and $6$ for the defender.



\paragraph{LM~\cite{chen2024RDC}}
Rather than denoising, LM purifies adversarial examples by directly optimizing the input 
to maximize its likelihood under the diffusion model, formulated as:
\begin{equation}
\min_{\hat{x}} \mathbb{E}_{\epsilon, t} \left\| \epsilon - \epsilon_\theta(\hat{x}_t, t) 
\right\|_2^2, \quad \text{s.t. } \| \hat{x} - x_0 \|_\infty \leq \eta,
\end{equation}
where $\hat{x}_t = \sqrt{\bar{\alpha}_{t'}} \hat{x} + \sqrt{1 - \bar{\alpha}_{t'}} \epsilon$ 
and $\eta$ is a predefined threshold. This is solved via projected gradient descent with a 
step size of $0.1$ over $5$ iterations, with $t$ sampled uniformly from $[0.4, 0.6]$. As 
we are concerned with adversarial purification, we only evaluate the LM purification step 
and not the full RDC pipeline.

\paragraph{ADBM~\cite{li2025adbmadversarialdiffusionbridge}}
ADBM improves upon DiffPure by learning a dedicated reverse bridge from the diffused adversarial distribution back to the clean distribution. Given $x_0$, an adversarial example is constructed as $x_0^a = x_0 + \epsilon_a$, which is then diffused to obtain $x_t^a$. The model learns to remove both Gaussian and adversarial noise through a modified forward process:
\[
x_t^d = \sqrt{\bar{\alpha}_t} x_0 + \sqrt{1 - \bar{\alpha}_t} \epsilon + \frac{\bar{\alpha}_T (1 - \bar{\alpha}_t)}{\sqrt{\bar{\alpha}_t}(1 - \bar{\alpha}_T)} \epsilon_a,
\]
and is trained using the following objective:
\begin{equation}
\mathcal{L}_b = \mathbb{E}_{\epsilon, t} \left\| \frac{\bar{\alpha}_T \sqrt{1 - \bar{\alpha}_t}}{(1 - \bar{\alpha}_T)\sqrt{\bar{\alpha}_t}} \epsilon_a + \epsilon - \epsilon_\theta(x_t^d, t) \right\|^2.
\end{equation}

The adversarial noise $\epsilon_a$ is generated by running PGD on the purification pipeline with a frozen classifier to maximize the classification loss after purification. To reduce training cost, the timestep $t$ and noise $\epsilon$ are fixed within each optimization step. This loss and adversarial noise generation is then used to fine-tune a pretrained Diffusion Model.

We fine-tune a pre-trained diffusion model using the configuration and implementation from the original paper: 30K training iterations, a maximum timestep of $200$, and PGD with 3 steps and a step size of $8/255$ to compute $\epsilon_a$. During inference, ADBM is equivalent to DiffPure but uses the learned model, with $t^* = 100$ and the DDIM reverse sampler given by:
\begin{equation}
\footnotesize
x_{t-1} = \sqrt{\bar{\alpha}_{t-1}} \left( \frac{x_t - \sqrt{1 - \bar{\alpha}_t} \epsilon_\theta(x_t, t)}{\sqrt{\bar{\alpha}_t}} \right) + \sqrt{1 - \bar{\alpha}_{t-1}} \cdot \epsilon_\theta(x_t, t).
\end{equation}

Both attacker and defender use a fixed time interval of $20$.

\paragraph{BEYOND~\cite{beyond}} 

BEYOND detects adversarial examples by leveraging self-supervised learning (SSL) to assess the consistency between an input image and its augmented variants (neighbors). It relies on two detection signals: \textit{label consistency} and \textit{representation similarity}. Given an SSL encoder $f$, a classification head $g$, and a projection head $h$, BEYOND first produces $k$ augmentations $\{\hat{x}_i\}_{i=1}^k$ of input $x$.
An input is flagged as adversarial if the number of neighbors sharing the same label with $x$ falls below a threshold $T_{\text{label}}$, or if the cosine similarity between $r(x)$ and $r(\hat{x}_i)$ is below $T_{\text{cos}}$ for too many neighbors.
We use $k = 50$ neighbors, generated using augmentations (e.g., cropping, flipping, color jitter). We use the same SSL models trained with SimSiam from Solo-learn~\cite{sololearn}, but implement the detection procedure ourselves based on the pseudocode provided in the original paper.

\subsection{Attack configurations}

\paragraph{PGD~\cite{madry2018towards}}
For PGD, we use $\epsilon=8/255$, $\alpha=2/255$ and $n=10$, unless otherwise specified.

\paragraph{CW~\cite{cw2016attack}}
For CW, we use the implementation of FoolBox~\cite{rauber2017foolbox}. We perform $9$ binary search steps for $c$ starting at $0.001$. We use $50$ steps of size $0.01$. Confidence is set to $0$.

\paragraph{BPDA~\cite{athalye2018obfuscatedgradientsfalsesense}}
For BPDA, we use the implementation provided by Wang et al.~\cite{wang2024diffhammer}. The parameters are $\epsilon=8/255$, $\alpha=0.007$ and $n=50$.

\paragraph{DH~\cite{wang2024diffhammer}}
For DH, we use the default settings. These are $\epsilon=8/255$, $\alpha=0.007$ and $n=50$. We use $5$ EM samples. DH is allowed up to $3$ restarts per input, restarting for samples with a lower attack success rate than $50\%$ across $10$ purification runs.

\subsection{FlowPure implementation details}
\label{sec:add-fp}

\paragraph{\textbf{Training}.}
$\text{FlowPure}^{\text{PGD}}$ is trained on PGD samples with $\epsilon \sim \mathcal{U}(0, 0.05)$, fixing remaining parameters to $n=10$ and $\alpha=2/255$. $\text{FlowPure}^{\text{CW}}$ is trained on CW samples with randomized parameters: $n \sim \mathcal{U}(1, 50)$, $c \sim \mathcal{U}(0, 2)$, $\kappa \sim \mathcal{U}(0, 1)$, and $lr \sim \mathcal{U}(0, 0.2)$. In the interest of computational efficiency and diversity, we omit the hyperparameter search and do not stop optimization upon finding an adversarial example during training. These modifications are applied only during training; evaluation uses the unmodified CW attack.

$\text{FlowPure}^{\text{Gauss}}$ is trained as a standard generative CNF on clean data, using Gaussian noise $\varepsilon \sim \mathcal{N}(0, I)$ as the source distribution, with no adversarial examples required during training.

\paragraph{\textbf{Inference}.}
All FlowPure variants integrate the learned velocity field from $t=0$ to $t=1$ using Euler's method with $N=10$ steps:
\begin{equation}
    x_{t + \Delta t} \leftarrow x_t + v_\theta(t, x_t)\, \Delta t.
\end{equation}
The attacker's surrogate process uses $N=5$ steps. For $\text{FlowPure}^{\text{Gauss}}$, intermediate noise injection is applied at each Euler step as described in Algorithm~\ref{alg:stoch-flowpure} in the main paper.

\paragraph{\textbf{Derivation of $\sigma$.}}
Under OT paths, any intermediate sample can be written as $x_t = (1-t)x_0 + t x_1$, with $x_0 \sim \mathcal{N}(0,I)$. We seek $\sigma$ such that $\tilde{x} = \eta \cdot x_t + \sigma \cdot \varepsilon$ has the same marginal distribution as $x_{\eta t} = (1 - \eta t)x_0 + \eta t\, x_1$. Expanding $x_t$:
\[
\tilde{x} = \eta t\, x_1 + \eta(1-t) x_0 + \sigma \varepsilon.
\]
Since both $x_0$ and $\varepsilon$ are drawn from $\mathcal{N}(0,I)$, the noise term is Gaussian with variance $\eta^2(1-t)^2 + \sigma^2$. Matching this to the variance of $x_{\eta t}$, which is $(1 - \eta t)^2$, gives:
\[
\eta^2(1-t)^2 + \sigma^2 = (1 - \eta t)^2,
\]
and therefore:
\begin{equation}
    \sigma = \sqrt{(1 - \eta t)^2 - \eta^2(1-t)^2}.
\end{equation}

\section{Additional Experimental Results}
\label{app:addexpres}

In this section, we provide additional experimental results that complement and extend the main findings presented in the paper. These results offer further insight into the behavior, robustness, and generalization properties of FlowPure under a variety of settings and evaluation protocols.

\begin{figure*}[t!]
    \centering
    \includegraphics[width=1\linewidth]{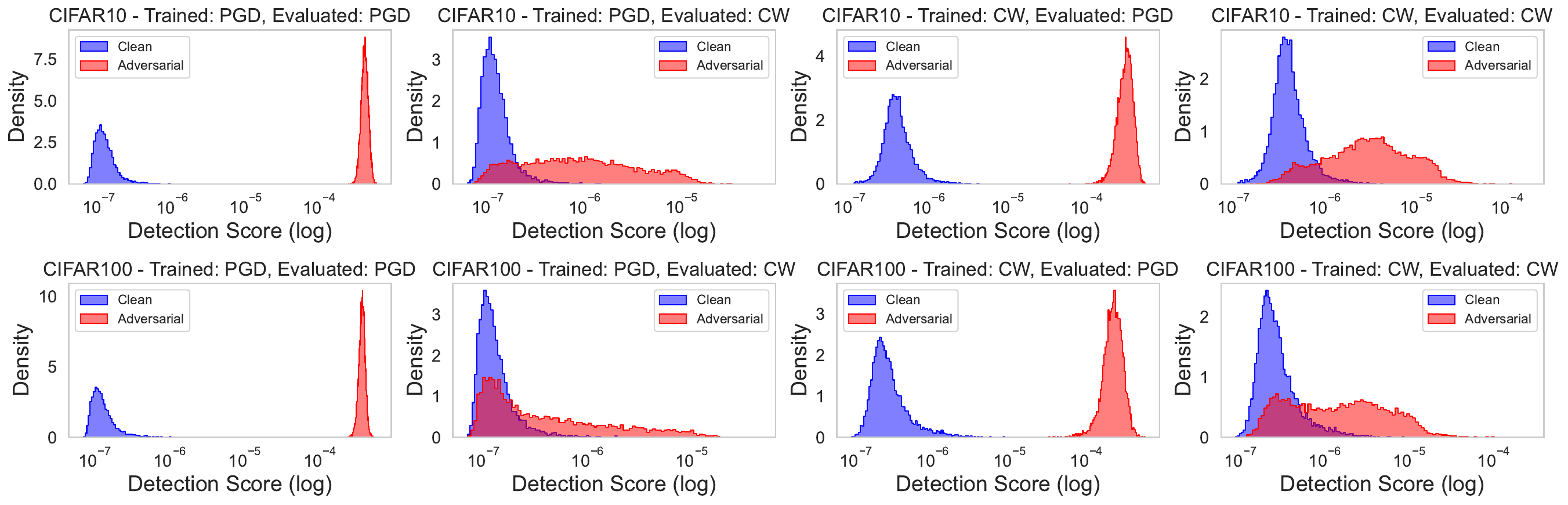}
    \caption{Distribution of detection scores for PGD and CW attacks. PGD attacks can be detected with near-perfect accuracy by FlowPure in both CIFAR-10 and CIFAR-100. }
    \label{fig:detection-scores}
\end{figure*}

\textbf{Detection}: Figure \ref{fig:detection-scores} displays the distribution of velocities used as detection scores for clean and adversarial samples. We consider an attack-aware scenario (detector is trained and evaluated in the same attack), and a transfer scenario (detector is trained and evaluated on different attacks). Regarding the attack-aware PGD detector, both distributions exhibit non-overlapping detection scores on both CIFAR-10 and CIFAR-100, enabling near-perfect detection of attacks. Moreover, this trend can be also observed with CW-trained detectors, as those seem to be efficient at detection of PGD adversarial samples. However, CW adversarial examples, both in the attack-aware and in the transfer scenario, demonstrate greater overlap in detection scores with clean samples, likely due to smaller perturbations of CW with less pronounced initial velocity field magnitudes, making them more challenging to distinguish from legitimate samples.

\textbf{Ablation:} Table~\ref{tab:wb-tradeoffs} reports the ablation study of Gaussian FlowPure under different noise configurations, and compares its performance to DiffPure in both preprocessor-blind and fully adaptive white-box scenarios on CIFAR-10 and CIFAR-100. 

\begin{table*}[tb!]
\centering
\scriptsize
\caption{Robustness of Gaussian FlowPure ($FP^{Gauss}$) and DiffPure\cite{diffpure} on CIFAR-10 and CIFAR-100, for various noise injection levels. For FlowPure, $\sigma$ denotes standard deviation of Gaussian noise injected before purification. For DiffPure, $t$ denotes how far along the forward process an image is diffused before purification.}
\begin{tabular}{@{}llccccccclll@{}}
\toprule
\multirow{3}{*}{} & \textbf{Method} & \multicolumn{2}{c}{\textbf{Std. Accuracy}} & \multicolumn{1}{l}{} & \multicolumn{7}{c}{\textbf{Robust Accuracy}} \\ \midrule 
 &  & \multicolumn{1}{l}{} & \multicolumn{1}{l}{} & \multicolumn{1}{l}{} & \multicolumn{4}{c}{\textbf{Preprocessor-blind}} &  & \multicolumn{2}{c}{\textbf{White-box}} \\ \cmidrule(lr){3-4} \cmidrule(lr){6-9} \cmidrule(l){11-12} 
 &  & $avg$ & $worst$ & \multicolumn{1}{l}{} & PGD$_{\text{avg}}$ & PGD$_{\text{worst}}$ & CW$_{\text{avg}}$ & CW$_{\text{worst}}$ &  & \multicolumn{1}{c}{$DH_{avg}$} & \multicolumn{1}{c}{$DH_{worst}$} \\ \midrule
\multirow{13}{*}{\rotatebox[origin=c]{90}{\textbf{CIFAR-10}}} & \textsl{Victim} & $95.81$ & $95.81$ &  & $0.04$ & $0.04$ & $0.00$ & $0.00$ &  & \multicolumn{1}{c}{--} & \multicolumn{1}{c}{--} \\
 & \textsl{DiffPure}: &  &  &  &  &  &  &  &  &  &  \\
 & \:\: $t=50$ & 93.51 & 83.18 & & 91.09 & 77.60 & 93.24 & 82.34 & & $17.83 \pm 0.61$ & $5.34 \pm 1.44$ \\
 & \:\: $t=80$ & 91.61 & 78.10 & & 90.42 & 75.32 & 91.31 & 76.96 & & $35.25 \pm 1.49$ & $13.93 \pm 1.58$ \\
 & \:\: $t=100$ & 89.97 & 73.81 & & 89.00 & 71.89 & 89.79 & 73.42 & & $39.74 \pm 1.59$ & $17.25 \pm 0.90$ \\
 & \:\: $t=130$ & 86.95 & 67.49 & & 86.46 & 66.48 & 87.03 & 67.40 & & $41.76 \pm 1.37$ & $19.27 \pm 1.26$ \\
 & \:\: $t=150$ & 85.14 & 63.30 & & 84.72 & 62.71 & 85.06 & 63.22 & & $41.74 \pm 1.04$ & $18.16 \pm 0.52$ \\
 & \textsl{$\text{FP}^{\text{Gauss}}$}: &  &  &  &  &  &  &  &  &  &  \\
 & \:\: $t=0.9$ & 92.65 & 81.89 & & 91.29 & 78.98 & 92.53 & 81.35 & & $32.44 \pm 0.87$ & $13.15 \pm 0.98$ \\
 & \:\: $t=0.875$ & 91.15 & 78.11 & & 90.33 & 76.24 & 91.04 & 77.84 & & $40.42 \pm 1.64$ & $19.08 \pm 0.92$ \\
 & \:\: $t=0.85$ & 89.60 & 74.59 & & 88.81 & 72.79 & 89.44 & 74.31 & & $43.12 \pm 1.49$ & $21.88 \pm 1.28$ \\
 & \:\: $t=0.825$ & 87.51 & 70.48 & & 87.06 & 69.35 & 87.39 & 69.78 & & $44.35 \pm 0.75$ & $22.72 \pm 0.98$ \\
 & \:\: $t=0.8$ & 85.20 & 65.22 & & 84.96 & 65.12 & 85.26 & 65.48 & & $43.50 \pm 1.02$ & $21.81 \pm 0.45$ \\ \midrule
\multirow{13}{*}{\rotatebox[origin=c]{90}{\textbf{CIFAR-100}}}& \textsl{Victim} & $80.73$ & $80.73$ &  & $0.01$ & $0.01$ & $0.00$ & $0.00$ &  & \multicolumn{1}{c}{--} & \multicolumn{1}{c}{--} \\
 & \textsl{DiffPure}: &  &  &  &  &  &  &  &  &  &  \\
 & \:\: $t=50$ & 71.39 & 49.40 & & 65.99 & 42.78 & 70.81 & 48.39 & & $6.33 \pm 0.59$ & $2.02 \pm 0.11$ \\
 & \:\: $t=80$ & 65.30 & 41.22 & & 62.74 & 38.57 & 64.97 & 41.00 & & $13.08 \pm 0.92$ & $4.10 \pm 1.53$ \\
 & \:\: $t=100$ & 61.62 & 37.54 & & 59.94 & 35.52 & 61.48 & 37.43 & & $15.73 \pm 0.59$ & $5.34 \pm 0.63$ \\
 & \:\: $t=130$ & 56.51 & 31.92 & & 55.63 & 31.08 & 56.56 & 32.24 & & $16.93 \pm 0.56$ & $5.27 \pm 1.01$ \\
 & \:\: $t=150$ & 53.46 & 29.19 & & 52.67 & 28.24 & 53.50 & 29.41 & & $16.82 \pm 0.39$ & $5.01 \pm 0.63$ \\
 & \textsl{$\text{FP}^{\text{Gauss}}$}: &  &  &  &  &  &  &  &  &  &  \\
 & \:\: $t=0.9$ & 68.41 & 46.76 & & 65.53 & 43.04 & 68.20 & 46.22 & & $10.72 \pm 0.71$ & $3.91 \pm 1.09$ \\
 & \:\: $t=0.875$ & 64.64 & 41.57 & & 62.91 & 39.58 & 64.34 & 41.31 & & $14.26 \pm 0.53$ & $4.95 \pm 0.69$ \\
 & \:\: $t=0.85$ & 60.75 & 37.31 & & 59.57 & 35.76 & 60.65 & 37.07 & & $16.06 \pm 0.30$ & $5.34 \pm 0.92$ \\
 & \:\: $t=0.825$ & 57.36 & 33.47 & & 56.22 & 32.32 & 56.75 & 33.27 & & $16.71 \pm 0.40$ & $5.27 \pm 0.85$ \\
 & \:\: $t=0.8$ & 53.45 & 29.62 & & 52.86 & 28.89 & 53.57 & 29.90 & & $16.73 \pm 0.26$ & $5.08 \pm 1.09$ \\ \bottomrule
 
\end{tabular}
\label{tab:wb-tradeoffs}

\end{table*}

\textbf{AutoAttack:} Tables~\ref{tab:aa_results_linf} and ~\ref{tab:aa_results_l2} present the complete results of our transferability analysis to previously unseen attacks and perturbation norms, evaluated using the individual components of the AutoAttack benchmark suite on CIFAR-10 and CIFAR-100. 

\begin{table*}[t!]
\caption{Robustness of the baselines and both variants of FlowPure (deterministic and Gaussian) against all attacks comprised in the AutoAttack benchmark ($L_\infty$ norm ), in the preprocessor-blind setting. The top performing method is shown in bold.}
\label{tab:aa_results_linf}
\scriptsize
\centering
\setlength{\tabcolsep}{5pt}
    \renewcommand{\arraystretch}{1.1}
\begin{tabular}{@{}llllllllll@{}}
\toprule
\multirow{3}{*}{} & \multicolumn{1}{c}{\multirow{3}{*}{\textbf{Method}}} & \multicolumn{8}{c}{\textbf{$L_{\infty}$}} \\ \cmidrule(l){3-10} 
 & \multicolumn{1}{c}{} & \multicolumn{2}{c}{$APGD$} & \multicolumn{2}{c}{$APGDT$} & \multicolumn{2}{c}{$FAB$} & \multicolumn{2}{c}{$Square$} \\ \cmidrule(l){3-10} 
 & \multicolumn{1}{c}{} & \multicolumn{1}{c}{avg} & \multicolumn{1}{c}{worst} & \multicolumn{1}{c}{avg} & \multicolumn{1}{c}{worst} & \multicolumn{1}{c}{avg} & \multicolumn{1}{c}{worst} & \multicolumn{1}{c}{avg} & \multicolumn{1}{c}{worst} \\ \midrule
\multirow{9}{*}{\rotatebox[origin=c]{90}{\textbf{CIFAR-10}}} & \textit{Baselines}: &  &  &  &  &  &  &  &  \\
 & \:\: DiffPure~\cite{diffpure} & 89.12 & 72.14 & 88.84 & 71.55 & 89.76 & 73.46 & 88.90 & 71.18 \\
 & \:\: GDMP~\cite{wang2022guided} & \underline{91.59} & 78.60 & \textbf{91.32} & 78.10 & 92.41 & 80.62. & \textbf{91.17} & \textbf{77.28} \\
 & \:\: LM~\cite{lee2023robustevaluationdiffusionbasedadversarial} & 66.91 & 43.38 & 67.69 & 44.05 & 77.87 & 54.99 & 77.38 & 55.65 \\
 & \:\: ADBM~\cite{li2025adbmadversarialdiffusionbridge} & 90.37 & 72.19 & \underline{90.05} & 71.32 & 89.87 & 73.64 & \underline{89.06} & 71.79 \\
 & \textit{Ours}: &  &  &  &  &  &  &  &  \\
 & \:\: FlowPure$^{\text{PGD}}$ & \multicolumn{1}{c}{\textbf{92.07}} & \multicolumn{1}{c}{\textbf{92.08}} & \multicolumn{1}{c}{81.79} & \multicolumn{1}{c}{\underline{81.80}} & \multicolumn{1}{c}{\underline{94.58}} & \multicolumn{1}{c}{\underline{94.58}} & \multicolumn{1}{c}{64.96} & \multicolumn{1}{c}{64.97} \\
 & \:\: FlowPure$^{\text{CW}}$ & \multicolumn{1}{c}{89.20} & \multicolumn{1}{c}{\underline{89.21}} & \multicolumn{1}{c}{87.87} & \multicolumn{1}{c}{\textbf{87.88}} & \multicolumn{1}{c}{\textbf{94.95}} & \multicolumn{1}{c}{\textbf{94.95}} & \multicolumn{1}{c}{67.74} & \multicolumn{1}{c}{67.75} \\
 & \:\: $\text{FlowPure}^{\text{Gauss}}$ & 88.91 & 73.31 & 88.82 & 73.02 & 89.41 & 74.21 & 88.48 & \underline{71.89} \\ \midrule
\multirow{9}{*}{\rotatebox[origin=c]{90}{\textbf{CIFAR-100}}} & \textit{Baselines}: &  &  &  &  &  &  &  &  \\
 & \:\: DiffPure~\cite{diffpure} & \underline{60.28} & 35.57 & \underline{60.14} & 35.69 & 61.63 & 37.71 & \underline{60.03} & \underline{35.40} \\
 & \:\: GDMP~\cite{wang2022guided} & \textbf{64.80} & \underline{41.89} & \textbf{64.43} & 41.32 & 66.01 & 43.67 & \textbf{63.61} & \textbf{39.96} \\
 & \:\: LM~\cite{lee2023robustevaluationdiffusionbasedadversarial} & 27.99 & 10.69 & 29.15 & 11.31 & 39.78 & 17.27 & 35.92 & 15.89 \\
 & \:\: ADBM~\cite{li2025adbmadversarialdiffusionbridge} & 60.19 & 28.57 & 58.92 & 28.00 & 58.82 & 30.69 & 57.00 & 28.18 \\
 & \textit{Ours}: &  &  &  &  &  &  &  &  \\
 & \:\: FlowPure$^{\text{PGD}}$ & \multicolumn{1}{c}{59.63} & \multicolumn{1}{c}{\textbf{59.63}} & \multicolumn{1}{c}{49.12} & \multicolumn{1}{c}{\textbf{49.12}} & \multicolumn{1}{c}{\textbf{78.81}} & \multicolumn{1}{c}{\textbf{78.81}} & \multicolumn{1}{c}{18.05} & \multicolumn{1}{c}{18.05} \\
 & \:\: FlowPure$^{\text{CW}}$ & \multicolumn{1}{c}{40.98} & \multicolumn{1}{c}{40.98} & \multicolumn{1}{c}{43.00} & \multicolumn{1}{c}{\underline{43.00}} & \multicolumn{1}{c}{\underline{78.65}} & \multicolumn{1}{c}{\underline{78.65}} & \multicolumn{1}{c}{32.01} & \multicolumn{1}{c}{32.01} \\
 & \:\: $\text{FlowPure}^{\text{Gauss}}$ & 59.80 & 35.75 & 59.55 & 35.95 & 60.56 & 37.26 & 59.42 & 35.37 \\ \bottomrule
\end{tabular}
\end{table*}

\begin{table*}[t!]
\caption{Robustness of the baselines and both variants of FlowPure (deterministic and Gaussian) against all attacks comprised in the AutoAttack benchmark ($L_2$ norm), in the preprocessor-blind setting. The top performing method is shown in bold.}
\label{tab:aa_results_l2}
\scriptsize
\centering
\setlength{\tabcolsep}{5pt}
    \renewcommand{\arraystretch}{1.1}
\begin{tabular}{@{}llllllllll@{}}
\toprule
\multirow{3}{*}{} & \multicolumn{1}{c}{\multirow{3}{*}{\textbf{Method}}} & \multicolumn{8}{c}{\textbf{$L_2$}} \\ \cmidrule(l){3-10} 
 & \multicolumn{1}{c}{} & \multicolumn{2}{c}{$APGD$} & \multicolumn{2}{c}{$APGDT$} & \multicolumn{2}{c}{$FAB$} & \multicolumn{2}{c}{$Square$} \\ \cmidrule(l){3-10} 
 & \multicolumn{1}{c}{} & \multicolumn{1}{c}{avg} & \multicolumn{1}{c}{worst} & \multicolumn{1}{c}{avg} & \multicolumn{1}{c}{worst} & \multicolumn{1}{c}{avg} & \multicolumn{1}{c}{worst} & \multicolumn{1}{c}{avg} & \multicolumn{1}{c}{worst} \\ \midrule
\multirow{9}{*}{\rotatebox[origin=c]{90}{\textbf{CIFAR-10}}} & \textit{Baselines}: &  &  &  &  &  &  &  &  \\
 & \:\: DiffPure~\cite{diffpure} & 89.25 & 72.60 & 89.13 & 72.08 & 89.76 & 73.46 & 89.63 & 72.96 \\
 & \:\: GDMP~\cite{wang2022guided} & \textbf{91.73} & \textbf{79.33} & \textbf{91.64} & \textbf{78.63} & \underline{92.40} & 80.60 & \textbf{92.07} & 79.90 \\
 & \:\: LM~\cite{lee2023robustevaluationdiffusionbasedadversarial} & 70.62 & 49.03 & 70.45 & 48.29 & 79.96 & 59.48 & 79.63 & 60.12 \\
 & \:\: ADBM~\cite{li2025adbmadversarialdiffusionbridge} & \underline{90.66} & \underline{75.80} & \underline{90.22} & \underline{74.80} & 91.65 & 77.51 & \underline{90.93} & 76.34 \\
 & \textit{Ours}: &  &  &  &  &  &  &  &  \\
 & \:\: FlowPure$^{\text{PGD}}$ & 16.40 & 16.40 & 10.00 & 10.00 & 91.84 & \underline{91.84} & 80.96 & \underline{80.96} \\
 & \:\: FlowPure$^{\text{CW}}$ & 51.84 & 51.48 & 46.66 & 46.66 & \textbf{93.77} & \textbf{93.77} & 85.34 & \textbf{85.34} \\
 & \:\: $\text{FlowPure}^{\text{Gauss}}$ & 89.07 & 73.71 & 89.14 & 73.45 & 89.41 & 74.22 & 89.12 & 73.17 \\ \midrule
\multirow{9}{*}{\rotatebox[origin=c]{90}{\textbf{CIFAR-100}}} & \textit{Baselines}: &  &  &  &  &  &  &  &  \\
 & \:\: DiffPure~\cite{diffpure} & \underline{60.84} & 36.35 & \underline{60.61} & 36.22 & 61.54 & 37.30 & 61.02 & 36.94 \\
 & \:\: GDMP~\cite{wang2022guided} & \textbf{65.06} & \textbf{42.35} & \textbf{64.91} & \textbf{42.33} & 66.08 & 43.63 & \textbf{65.68} & 43.04 \\
 & \:\: LM~\cite{lee2023robustevaluationdiffusionbasedadversarial} & 32.61 & 13.88 & 31.47 & 13.71 & 42.98 & 19.85 & 40.84 & 19.90 \\
 & \:\: ADBM~\cite{li2025adbmadversarialdiffusionbridge} & 61.01 & 33.65 & 60.52 & 32.67 & 63.22 & 35.56 & \underline{62.43} & 34.43 \\
 & \textit{Ours}: &  &  &  &  &  &  &  &  \\
 & \:\: FlowPure$^{\text{PGD}}$ & 4.47 & 4.47 & 5.07 & 5.07 & \underline{78.43} & \underline{78.43} & 43.88 & \underline{43.88} \\
 & \:\: FlowPure$^{\text{CW}}$ & 18.57 & 18.57 & 19.73 & 19.73 & \textbf{78.87} & \textbf{78.87} & 57.11 & \textbf{57.11} \\
 & \:\: $\text{FlowPure}^{\text{Gauss}}$ & 60.22 & \underline{36.47} & 59.92 & \underline{36.37} & 60.71 & 37.27 & 60.40 & 36.93 \\ \bottomrule
\end{tabular}
\end{table*}

\section{Computational Resources}
\label{app:compres}

 \textbf{Implementation.} All experiments were performed on PyTorch version 2.3.951, using a NVIDIA GeForce RTX 3090 with 24 GB of memory. The WideResNet classifiers were trained using a public implementation~\footnote{\url{https://github.com/bmsookim/wide-resnet.pytorch} (License: MIT)}. The diffusion models for CIFAR-10 and CIFAR-100 are trained using the implementation of Song et al.~\cite{song2021scorebased}. To apply CFM we use the TorchCFM library~\footnote{\url{https://github.com/atong01/conditional-flow-matching} (License: MIT)}, and to generate adversarial examples, we base our implementation on \textit{torchattacks}~\cite{kim2020torchattacks}. Training on each dataset for each adversarial attack is the same. Training is done on batches of size $64$ for $300000$ iterations, with the Adam optimizer with a learning rate of $0.0002$.

\textbf{Training.} The total training time of FlowPure (Table \ref{tab:comp_time_training}) is dominated by the computation of adversarial attacks, as we chose a naive implementation for simplicity, in which adversarial examples are computed inside the training loop. However, the adversarial examples used depend exclusively on the classifier, and as such, can be generated prior to training the CNF. Additionally, as we noticed some degree of transferability in our experiments, the use of computationally simpler attacks during training remains possible in theory.

\textbf{Inference.} During inference (shown in Table \ref{tab:comp_time_inf}),  computation time of each purification method is dominated by the number of times the diffusion or flow model is evaluated. Given that every method uses the same architecture, this is comparable for all considered methods. The observed differences can be attributed to the use of the default configuration of each approach, as the step size can then be tuned to make trade-offs between computation time and robustness. The only exception to this would be LM~\cite{chen2024RDC}, which requires computing gradients with respect to the diffusion model.

\begin{table}[ht!]
\centering
\caption{Training time per batch (64 samples) for PGD, CW, and Gaussian CNF, including contributions from gradient descent and noise generation.}
\scriptsize

\vspace{0pt} 
\centering
\begin{tabular}{lrrr}\toprule
& \textbf{PGD (s)} & \textbf{CW (s)} & \textbf{Gaussian (s)}\\\midrule
Total & $0.91 \pm 3.1\text{e}{-3}$  & $5.52 \pm 6.4\text{e}{-2}$ & $0.31 \pm 3.2\text{e}{-3}$ \\
Gradient Descent & $0.31 \pm 8.7\text{e}{-4}$ & $0.32 \pm 1.3\text{e}{-3}$ & $0.31 \pm 3.2\text{e}{-3}$ \\
Noise Generation & $0.60 \pm 2.7\text{e}{-3}$ & $5.20 \pm 6.4\text{e}{-2}$ & $2.3\text{e}{-3} \pm 7.9\text{e}{-5}$ \\
\bottomrule
\end{tabular}
\label{tab:comp_time_training}
\end{table}

\begin{table}[ht!]
\scriptsize
\hspace{1.5cm}
\vspace{0pt} 
\centering
\caption{Combined purification and classification time per batch (64 samples) during inference for each purifier.}
\begin{tabular}{lr}\toprule
& \textbf{Inference Time (s)} \\ \midrule
Victim     & $0.01 \pm 5.4\text{e}{-3}$ \\ \midrule
DiffPure    & $1.33 \pm 3.5\text{e}{-3}$  \\
GDMP        & $3.18 \pm 1.2\text{e}{-2}$ \\
LM          & $1.74 \pm 6.0\text{e}{-3}$  \\
ADBM        & $0.67 \pm 2.4\text{e}{-3}$  \\
FlowPure    & $1.20 \pm 4.0\text{e}{-3}$  \\
\bottomrule
\end{tabular}


\label{tab:comp_time_inf}
\end{table}

\end{document}